\documentclass[12pt, oneside]{book}
\usepackage[left=1.5in, top=1in, right=1in, bottom=1in]{geometry}
\usepackage{cite}
\usepackage[doublespacing]{setspace}
\usepackage{amsmath}
\usepackage{amsthm}
\usepackage{amsfonts}
\usepackage{mathtools}
\usepackage{booktabs} 
\usepackage[numbers]{natbib}
\newtheorem{prop}{Proposition}
\newtheorem{theorem}{Theorem}
\theoremstyle{definition}
\newtheorem{defn}[theorem]{Definition}
\begin{document}
\begin{titlepage}

\begin{singlespace}
\begin{titlepage}

\begin{center}

{\centering\fontsize{12}{12}
\uppercase{Motion Equivariance of Event-based Camera Data with the Temporal Normalization Transform}
}
\vspace{48pt}

Ziyun Wang

\vspace{48pt}

A THESIS

\vspace{24pt}

in

\vspace{24pt}

Robotics

\vspace{48pt}

Presented to the Faculties of the University of Pennsylvania in Partial
Fulfillment of the Requirements for the Degree of Master of Science\\

\vspace{24pt}

2019

\end{center}
\vspace{48pt}
\noindent\rule{180pt}{0.4pt}

\noindent Kostas Daniilidis

\noindent Supervisor of Thesis

\vspace{36pt}
\noindent\rule{180pt}{0.4pt}

\noindent Camillo J. Taylor

\noindent Graduate Group Chairperson
\end{titlepage}
\end{singlespace}
\end{titlepage}
\frontmatter
\chapter{Dedication}

This thesis is dedicated to the memory of my grandfather. He had been a quiet inspiration and my strongest support for my decision to purse a graduate degree.

\chapter*{Acknowledgments}
I am grateful to Dr. Kostas Daniilids for providing me the opportunities to work in his group. He has provided valuable advice in both my research and my personal development.\\\\
\noindent I want to thank Alex Zihao Zhu, for collaborating with me in this project. His experience and knowledge have benefited me tremendously in completing the thesis. \\\\

\tableofcontents
\listoffigures
\listoftables
\mainmatter
\chapter*{Abstract}
The event-based camera is a bio-inspired camera that captures the change of logarithmic light intensity of an image. As opposed to integrating the light information within certain time intervals and forming synchronous frames for traditional cameras, the event-based camera asynchronously generates events with their corresponding pixel locations, timestamps and polarities. Due to the unique design of the novel camera sensor, it has advantages over the traditional camera in many areas, such as higher dynamic range, lower latency and reduced motion blur. However, the novel camera also has some major differences from the frame-based cameras that make traditional computer vision algorithms less suitable for the event camera data. For example, compared with the traditional cameras, the event cameras capture the changes of the light intensities rather than the absolute photometric intensities of the pixels. Consequently, to work with event data, one needs to consider the dynamic motions in the scene. In addition, an important benefit of the event camera is its ability to preserve the valuable temporal information, and we want to fully incorporate the information into our algorithms. With these differences, new computer vision algorithms need to be developed to handle event based camera data effectively.
\\\\
In this work, we focus on using convolution neural networks (CNN) to perform object recognition on the event data. In object recognition, it is important for a neural network to be robust to the variations of the data during testing. For traditional cameras, translations are well handled because CNNs are naturally equivariant to translations. However, because event cameras record the change of light intensity of an image, the geometric shape of event volumes will not only depend on the objects but also on their relative motions with respect to the camera. The deformation of the events caused by motions causes the CNN to be less robust to unseen motions during inference. To address this problem, we would like to explore the equivariance property of CNNs, a well-studied area that demonstrates to produce predictable deformation of features under certain transformations of the input image.
\\\\
In this thesis project, we explore the motion equivariance properties of event-based camera data to convolution. We explore some background knowledge of event-based cameras and the mathematical basis for equivariance. In addition, we include our recent project on applying the equivariance theory to improving the robustness of computer vision algorithms in object recognition for event data. A novel coordinate transform is proposed to make CNNs equivariant to global motions. 

\chapter{Event-based Camera}
\section{A Bio-inspired Vision Sensor}
With the recent development of computer vision and deep learning, algorithms using traditional frame-based camera images have achieved considerable success in many applications. However, despite the final output of cameras matching what we see in our minds, the way cameras construct pictures through accumulating photons on a optical sensor is very different from the human vision system. Human eyes have cells that perceive light independently and let the information propagate through the optic nerve to the visual cortex of the brain for asynchronous processing. 
\\\\
In order to push the boundaries of machine intelligence to human levels, attempts have been made to reproduce how human-beings perceive the world in hardware designs. In this pursuit, event cameras are designed to mimic human eyes by tracking the relative change of light in a scene asynchronously. With the novel design of the camera, how to use its unique output effectively becomes an important field of research. In this chapter, we will cover how the events are generated and how they are used in modern computer vision research. In addition, we will compare the event cameras with traditional cameras and their respective advantages and disadvantages.

\section{Traditional Frame-based Camera}
\label{cam-lim}

Traditional cameras mainly use frame-based architectures, where the light within a certain time interval is captured using either CMOS (Complementary metal–oxide–semiconductor) or CCDs (Charge Coupled Device). The synchronous outputs of the camera are interpreted as either single-channel or multi-channel images. These kinds of images are extensively studied and used in computer vision. However, traditional cameras have several limitations due to their frame-based design:

\subsection{Dynamic Range}
Traditional cameras tend to have very limited dynamic range. Because of the frame-based design, all image pixels are exposed to light for the same amount of time, which gives the camera a very limited dynamic range. In natural environments, limited dynamic range can cause many problems in computer vision applications. For instance, let us assume there is a self-driving car operating in a dark environment, such as a tunnel. We want the perception system of the car to be able to see the objects well in the dark while avoiding being blinded by the strong headlights of another vehicle. With a limited dynamic range, it is difficult to accomplish theses two goals simultaneously due to the inherent limitations of the camera sensor.

\subsection{Latency and Bandwidth}
It is desirable for a camera to have high frame rate and low latency. Higher frame rate gives a better approximate of the continuous light signal and therefore is preferred in computer vision algorithms. For traditional frame-based cameras, the bandwidth required for high-speed photography is extremely high because all pixels need to be processed and transmitted synchronously. According to \cite{kleinfelder200110000}, the bandwidth required for processing a sequence of 352x288 pixel images at 10,000 frames per second requires 1.34 GB/s of bandwidth. This leads to extra cost in designing and manufacturing the hardware to deal with the large amount of camera data. Additionally, there exists redundancy in the frame-based camera data. For example, the pixels of a static object will be redundant after the initial frame in a sequence of images. There are techniques that focus on reducing the redundant pixels by using region-of-interest readout, but very complex control strategies are needed \cite{lichtsteiner}. Despite these efforts to make the camera faster, it is difficult to achieve extremely low latency with traditional cameras.

\subsection{Motion Blur}
Combating motion blur has been a challenging problem in high speed computer vision. Due to the frame-based design of the traditional cameras, the light signal within a certain amount of time is accumulated onto the same image. The earlier light within the time window is then blurred out by the later light. As the amount of movement increases in a scene, the amount of motion blur increases as well. Typically, a moving object will have more blurry look than a static object. Naturally, the problem becomes more obvious with low frame rate cameras, where the amount of information collapsed on the same frame is high. Motion blur makes tasks such as object recognition and classification more challenging.

\section{Advantages}
\label{event-char}
The unique design of the event camera gives it several advantages over traditional cameras in dynamic range, latency and motion blur.
\begin{itemize}
\item High Dynamic range: Event-based camera has much higher dynamic range. Note that in formula \ref{eq:threshold}, the left hand size is computed locally on each individual pixel location. Thus, there is no global constant pixel gain that limits the maximum range of light to be captured. Additionally, the designed circuit allows the camera to operate at extreme low light conditions due to its low threshold comparator. On the hand, the camera claims to operate in extremely high illumination conditions too. The DVS is reported to operate in the illumination range of 0.1 Lux to 100 kLux and the dynamic range is 120 dB, compared to roughly 60 dB for traditional cameras \cite{lichtsteiner}. 
\item Low Latency: Event-based cameras have much lower latency than frame-based cameras. As discussed in \ref{cam-lim}, a camera with 10k fps is regarded very high-speed and requires significant computational power and bandwidth to process the data. The temporal resolution of a 10kfps camera is $1 / 10kfps = 10^{-4}s$. In comparison, the response time of an event-based camera is $15\mu s$ or $1.5\times10^{-5}s$ \cite{lichtsteiner}. Furthermore, the temporal resolution of events is also much higher at $1\mu s$ because of the asynchronous design. It gives much finer discretization of the light signal than a frame-based camera. On the other hand, despite the high throughput of the camera, the event is triggered and transmitted individually without waiting in batch. Therefore, the data are no longer fighting for the bandwidth, making it much cheaper to design affordable on-board hardware for the sensor.

\item Reduced Motion blur: Given all events exist in x-y-time coordinate space, the events at the same pixel location are not compressed to one point in time. Rather, we preserve the valuable temporal information lost in frame-based camera data. The complete knowledge of the temporal change is very important in tasks like motion recognition and tracking. Additionally, the events give a more direct representation of the motions in the scene because we only consider the relative change in light intensity.

\end{itemize}
The above-mentioned benefits, in addition to features like low power consumption, make the event-based camera ideal in tasks such as fast feature tracking and visual odometry.

\section{Disadvantages}
While event cameras have shown advantages in many fields, it is not broadly used today. First, there are a lot fewer event cameras than traditional cameras on the market. As a result, we have much fewer event data sets than traditional image datasets. This makes it hard for researchers to explore different areas of application. Second, its novel design comes with a cost in many aspects. The handling and processing of asynchronous events requires re-thinking images from a different perspective. In this section, we will mainly discuss three of these challenges.

\subsection{Data Association}The event data are returned asynchronously without information of correspondence. It is challenging to infer whether an event is associated with an event that happened before or caused by a new feature in the image. If the association is known, we will be able to easily estimate the optical flow of the image and de-blur the events into a 2D pattern. However, such correspondence is not trivial to learn. The events are essentially points in the x-y-time space. Classical methods like Iterative Closest Points (ICP) are used to deal with the alignment/association problem for point clouds, but the efficiency and initialization quality are all important factors to consider. In traditional images, we can effectively use existing feature extractors, such as Scale-invariant feature transform (SIFT) \cite{lowe2004distinctive} and Histogram of oriented gradients (HOG) \cite{dalal2005histograms}, to obtain reliable features. Efforts are being made to to design generalized feature descriptor for events. New loss functions are also proposed to learn the correspondence automatically \cite{gallego2018unifying}. In the meantime, how to associate events remains a challenging problem.
 
\subsection{Asynchronous Processing} Traditional computer vision techniques mostly work with synchronous data frames. We assume everything is happening at the same time. With the event camera, we have two options: accumulate events as frames or process it asynchronously. The first option is less ideal, since we are not taking full advantage of camera's desirable low latency. On-line learning of asynchronous events has been an active field of study in the event camera research. There are works that explore the asynchronous nature of spiking neural networks to process events \cite{lee2016training, zhao2015feedforward, orchard2015hfirst, perez2013mapping}. However, there is a lack of hardware support for such architectures in real applications.

\subsection{Lack of Intensity Information} One thing missing from the event camera output is the intensity information of a pixel. The camera only generates uniform events (with polarity), which makes feature matching a more difficult task. Although the accumulative events can give us some information on intensity, this process also involves conversion of asynchronous events to some synchronous representation. Work has been done to reconstruct intensity information in 2D or 3D with event streams \cite{gehrig2018asynchronous, cook2011interacting, kim2008simultaneous, kim2016real}. However, these methods mostly depend on known environments or directly obtain the pixel intensities from the grayscale images.

\section{How Events are Generated}
\label{event_def}
\begin{figure}
\centering
\includegraphics[width=0.49\linewidth]{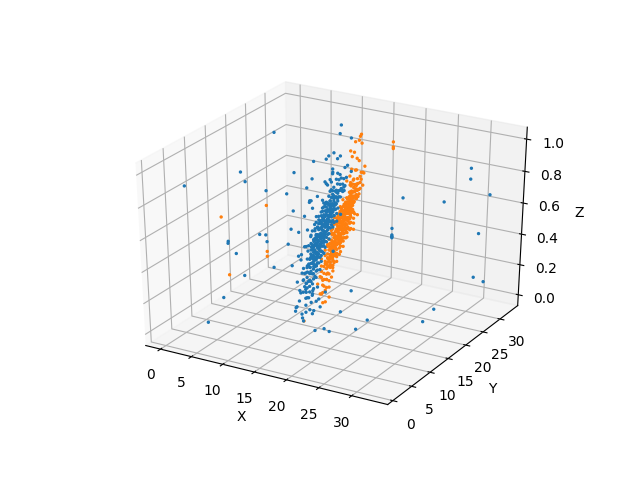}
\includegraphics[width=0.49\linewidth]{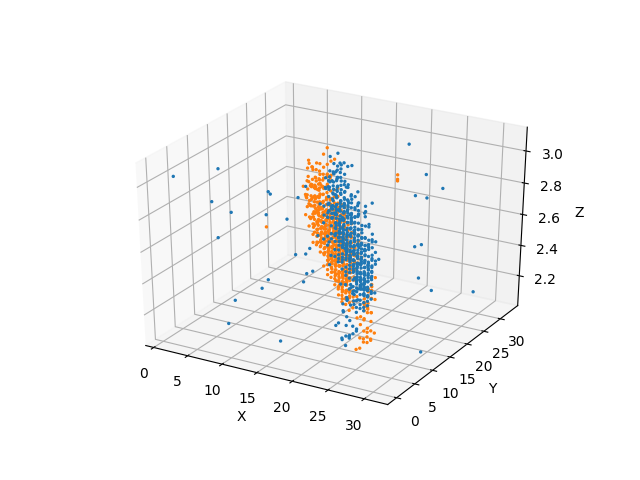}
\caption{Left: Digit moving in the positive x and positive y direction. Right: Digit moving in the negative x and negative y direction. Blue and orange points represent positive and negative events, respectively.}
\label{fig:event}
\end{figure}

This section describes how event-based camera data are generated through the Dynamic Vision Sensor (DVS) introduced in \cite{lichtsteiner}. First, we formally define temporal contrast ($TCON$) as: 
\begin{align}
TCON = \frac{1}{I(t)}\frac{dI(t)}{dt}=\frac{d(ln(I(t))}{dt}
\end{align} 
This term defines the change of intensity of pixels normalized by their current intensities. Intuitively, we want to measure how much a pixel has changed relative to its past value. Here $I(t)$ denotes the photocurrent at a particular timestamp, indicating the intensity of a pixel measured by the sensor. We integrate $TCON$ within a short amount of time to get a more robust estimate of such change. The integration process is achieved with a circuit designed in \cite{lichtsteiner}:
\begin{align}\Delta V_{diff} = -A \frac{U_T\kappa_{sf}}{\kappa_{fb}} \int_{t}^{t+\Delta t} TCON(t')dt'
\end{align}
Here $\Delta V_{diff}$ represents the accumulative sum of $TCON$ scaled by the hardware constants. Hyper-parameters $(M_{ONn}, M_{ONp}, M_{OFFn}, M_{OFFp})$ are chosen to determine if an event will be generated from a pixel position. When an event is generated, it also produces a polarity that indicates the direction of change of the logarithmic light intensity at the pixel. From another perspective, in the image domain, we replace the continuous current intensity function with a discretized image function. The trigger condition of an event is defined as:
\begin{align}
log(I(x, y, t_1)) - log(I(x, y, t_0)) > \epsilon \label{eq:threshold}
\end{align}
where $\epsilon$ is a hyperparameter threshold and $I(t)$ represents the image intensity. Such condition is satisfied when there is a sufficiently large change of light intensity at a pixel caused by its movement. In images, such movement is observed as the optical flow of the pixels. Generally, a pixel with high optical flow and high image gradient is more likely to trigger events. The polarity of an event indicates whether a pixel gets brighter or darker, which can be jointly determined by the direction of the optical flow and the image gradient of the triggering pixel.\\\\
A visualization of the events generated with the digit "1" with two different motions can be found in Figure \ref{fig:event}. The two sets of events are generated by the same number with two different motions. It can be seen that the shape of the events depends not only on the scene but also the motion. The positive events and negative events are generated on two different sides of the same stroke because the direction of the image gradients on the two sides are different with respect to the motion.

\section{Event Representation in Neural Networks}
\label{event-represent}

As described in the previous section, an event is parameterized by four values: $\mathbf{x} = (x, y)$ indicating the spatial position of an event in the camera plane, $\mathbf{t}$ the timestamp of the event and $\mathbf{p}$ the polarity of the event. We denote such  an event as $e(x, y, t, p)$. While this parameterization of an event is clear, there is no single established way to represent events as input data to a neural network. 
\\\\Naively, we can keep the raw output format of events from the camera, a $N \times 4$ matrix with each row as an event. While this keeps the complete information of the events, the convolution layers can no longer access the neighbors of the input pixels in direction computation. Consequentially, the neural network losses the direct spatial relationships within the input data as in the case of 2D images. In later sections, we will explore in detail why having spatial information (consequently translation equivariance) is critical in deep learning architectures.
\\\\Alternatively, we define an ``event image" that contains the accumulative number of events at each pixel, similarly to \cite{nguyen2017real}. This representation contains the 2D spatial knowledge of the events but discards the important information in the temporal domain, which is the main selling point of event-based cameras. \citet{zhu2017event} proposes a new representation that adds the most recent timestamps of the events at each pixel as a separate channel. Then the timestamped images are normalized by the size of the temporal window. This method preserves the temporal information to some degree and has shown to have achieved good performance in tasks such as flow estimation and visual odometry.
\\\\In our project described in Chapter \ref{content}, we aim to preserve the complete temporal and spatial information of the events while facilitating the training process of the deep neural network. Since discretized convolution is commonly adopted in modern deep learning architectures and frameworks, it is required to have some form of discretization of the events into a finite set of temporal bins. Additionally, the high resolution of the events in the temporal domain requires us to discretize the timestamps in a very fine way. Therefore, we adopt the event volume representation proposed by~\cite{zhu2018unsupervised}, inserting the events into an event volume weighted by the normalized timestamp in an interpolative way. Here is the formal definition of the approach:
\\\\For a set of $N$ events, $\{(x_i, y_i, t_i, p_i), i\in [1, N]\}$, the time dimension is discretized along $B$ bins. The timestamps of the events is then scaled to the range $[0, B-1]$, and the event volume is defined as:
\begin{align}
t^*_i =& (B-1)(t_i - t_1) / (t_{N} - t_1)\\
V(x,y,t)=&\sum_{i} p_i k_b(x-x_i)k_b(y-y_i)k_b(t-t^*_i)\\
k_b(a) =& \max(0, 1-|a|)
\end{align}
where $k_b(a)$ is the linear sampling kernel defined in \cite{jacobsen2017dynamic}. There are two main advantages of such representation: First, we preserve the complete spatiotemporal information of the events. Second, the events are inserted into a fixed-sized volume with interpolation without explicit rounding. This means the volume will retain the original distribution of the events parameterized by the pixel value of the event image. The function that maps the events to the even volume is fully differentiable, which allows the network to learn more effectively. 

\section{Motion of Events}

\subsection{Constant Flow Assumption}
\label{event-flow}
Constant local optical flow is a commonly used assumption in classical computer vision works such as \cite{lucas1981iterative}. Due to the high temporal resolution of the events, we are able to split an event stream into small temporal segments. Due to the short time interval of these segments, the motion within the time window can be assumed to be constant. In this work, we base our method on the constant flow assumption. However, this assumption is not strictly valid when we have bigger temporal windows or non-rigid objects. Such cases are left to be addressed in future work.

\subsection{Motion Model}
\label{event-motion}
In order to understand the effect of a motion, we need to formally formulate the motion model of events. If we assume an event $e(x, y, t, p)$ is generated by some feature parameterized by $(x_f, y_f)$. Imagine we have some event-triggering feature moving along the image space with constant optical flow. Section \ref{event-flow} assumes constant flow $(\dot{x}, \dot{y})$ within a short amount of time $\delta t$. The generated events are simply shifted by $(\dot{x} * \delta t_i, \dot{y} * \delta t_i)$ spatially in the x-y domain, where $\delta t_i$ is the duration of the motion. In our work, we set the timestamp of the oldest event in an event stream to be 0. Therefore, $\delta t$ simply becomes the zeroed timestamp of each event $t_i$. We model this transformation under constant optical flow as $L_{OF}$:
\begin{align*}
L_{OF} \begin{pmatrix}
\mathbf{x_i}\\t_i
\end{pmatrix} = 
\begin{pmatrix}
\mathbf{x_i} + \mathbf{\dot{x}}t_i\\t_i
\end{pmatrix}
\end{align*}
\noindent Note that the motion model is only valid on the assumption of constant motions of rigid objects. In our work, a small time window is selected in our work so the assumption holds reasonable. However, to deal with more complex motions (such as in the case of longer time window) or non-rigid objects, new factors need to be considered. We leave these more complex scenarios to our future work.

\chapter{Equivariance}
\section{Introduction}
With the recent success of Convolutional Neural Networks (CNN) in computer vision, we discover the translation equivariance property of the convolution layers (proof in \eqref{eq:trans}) plays an important role in its effectiveness \cite{cohen2016group, cohen2018spherical}. The translation symmetry of convolution makes it possible for kernels to share weights in different parts of the image, making the network more efficient in using its capacity in learning compared to the traditional fully connected layers. The translation symmetry is preserved through a deep CNN because of the translation equivariance property of convolution \cite{cohen2016group}. The equivariance property of the CNN can be defined as below: transformation of the image yields a homogeneous transformation of the feature maps after the convolution. In many tasks, it is particularly desirable to have such property. For example, in localization, semantic segmentation and optical flow estimation, the translated input image should naturally correspond to a translated output. More generally, it gives the features a predictable deformation after each convolutional layer. This means the spatial structure of the features will remain the same when the image is translated. The preserved spatial information can be potentially used in later prediction. With these benefits, it is very useful to achieve equivariance of CNN under certain transformations. In the next sections, we will explore some basic concepts and theoretical tools commonly used in the study of equivariance.
\section{Related Work}
\label{sec:equi-related}
Equivariance for CNNs is a well studied topic, and has roots in the study of Lie generators~\cite{ferraro1988relationship, segman1992canonical} and steerability~\cite{freeman1991design, simoncelli1992shiftable, teo1998design}. Recent works have extended these ideas for equivariance of CNNs to a number of transformations. \cite{cohen16_steer_cnns, jacobsen2017dynamic} combine steerability with neural networks. Harmonic Networks~\cite{worrall16_harmon_networ} use the complex harmonics to generate filters that are equivariant to both rotation and translation. \cite{cohen2016group} propose group convolutions, which performs convolutions using a group operation rather than translation.  More recently, \cite{cohen2018spherical, esteves2018learning} propose spherical representations of a 3D input, which are processed with convolutions on SO3 and spherical convolutions, respectively. Similar to this work, Polar Transform Networks~\cite{esteves} convert an image into its log polar form to gain equivariance to rotation and scaling, while obtaining translation invariance through a spatial transformer network. 

\section{Group Convolution}
First, we introduce the general form our commonly studied convolution: \textit{group-convolution}, as defined by \citet{kyatkin2000algorithms}. Let $f(x)$ and $\phi(x)$ be real valued functions on a transformation group G with $L_hf(g) = f(h^{-1}g)$, the group-convolution is defined as:
\begin{align}
(f \ast_G \phi)(x) = \int_{h\in G}f(h)\phi(h^{-1}g) dh
\end{align}
 In CNNs, a translational convolution is typically performed on the input image.
Therefore, we use the group action of $\mathbb{R}^2$ as the operator for the convolution, which is simply addition. The translational convolution can be written as:
\begin{align}
(f \ast \phi)(x) &= \int_{h\in \mathbb{R}^2}f(h)\phi(h^{-1}x) dh \\
& = \int_{h\in \mathbb{R}^2}f(h)\phi(x - h) dh
\end{align}
Note that while convolution can be performed on groups, translational convolution is commonly used in modern architectures. \citet{esteves} points  out that the evaluation of group convolution requires an appropriate measure $dh$ in order to perform integration.
\section{Group Equivariance and Convolution}
\begin{defn} Group Equivariance: A function $f$ is said to be equivariant under some transformation group $\mathbf{G}$ if for all $g \in \mathbf{G}$, where $L_g$ denotes a group action. A mapping $\Phi: E \to F$ is said to be equivariant to the group action $L_g$, $g \in G$ if: 
\begin{align}
\Phi(L_gf) =& L_g'(\Phi(f))
\end{align}
where $\L_g$ and $L_g'$ correspond to application of $g$ to $E$ and $F$ and satisfy $L_{gh} = L_gL_h$. In our case, we study the case when $\Phi$ is a mapping from the image to the features through convolution.
\end{defn}

\noindent The equivariance condition that needs to be satisfied for a group convolution is: For a convolution under some group $G$, in domain $x \in \mathbf{X}$. The group convolution $\ast_G : E \to F$ is said to be equivariant to the group action $L_a$, $a \in A$ if:
\begin{align}
((L_af)\ast_G\phi)(x)=&L_a((f\ast_G\phi))(x)
\end{align}
where $L_a$ corresponds to applying a group action $a$ to a function.
It is also proven that any convolution is equivariant under group convolution \cite{esteves}:
\begin{proof}
Letting $f(g)$ and $\psi(g)$ be real valued functions on G with $L_hf(g) = f(h^{-1}g)$
\begin{align}
(L_a f \ast_G \phi)(g) &= \int_{h\in G}f(a^{-1}h)\phi(h^{-1}g) dh \\
&= \int_{b\in G}f(b)\phi((ab)^{-1}g) db \\
&= \int_{b\in G}f(b)\phi((b^{-1}a^{-1}g) db \\
&= L_a ((f \ast_G \phi))(g) \end{align}
In other words, any group-convolution is always group equivariant. This implies that any group action on the image space will produce a homogeneous transformation on the feature space if our convolution is performed on the same group.
\end{proof}
\noindent This might sound counter-intuitive: \textit{if all group-convolutions are equivariant, why don't we  these convolutions in our architectures}? As pointed out in \cite{esteves}, this group equivariance property requires an appropriate measure $dh$ in order to perform integration, which is hard to obtain in most cases. While it is technically possible to implement group convolution via carefully designed methods and custom deep learning toolboxes, there is far less support for it than the standard translational convolution. In practice, we would like our method to be more general to well developed deep learning architectures.
\section{Example Proofs}
To give an example of CNN's equivariance to translation, here is a brief proof to show that translational convolution is equivariant to translation on $\mathbb{R}^2$, we define a translation t and denote $L_t$ as the translation applied to a function. We define $L_t$ to translate the signal by $-t$. $f: \mathbb{R}^2 \to \mathbb{R}^2$ and $\phi$ a filter $\phi: \mathbb{R}^2 \to \mathbb{R}^2$:
\begin{proof}
\begin{align} \label{eq:trans}
(L_t f \ast \phi)(x) &= \int_{h\in \mathbb{R}^2}f(h-t)\phi(x - h) dh \\
\intertext{Applying the variable substitution: $h' = h - t$}
&= \int_{a\in \mathbb{R}^2}f(h')\phi(x - (h' + t)) dh' \\
&= \int_{a\in \mathbb{R}^2}f(h')\phi((x - t) - h') dh' \\
&= L_t ((f \ast \phi))(x) \end{align}
\noindent In other words, a translation on the input space will yield the same translation in the output of the convolution, independent of what kernel values are chosen.
\end{proof}
\noindent The proof shows that applying a translation on the input images will translate the feature maps after the convolution as well. Here, we notice that the translation of the features map $L_t$ is the same as the translation on the input images $L_t$. This is called "invariance", a special case of equivariance. 
\\\\
Additionally, it is important to note that translational convolution is not equivariant to every transformation group $G$. Here we show an example of such case with a $SO(2)$ rotational group.
we define a 2D rotation r around the origin and denote $L_r$ as the translation applied to a function. $f$ is the mapping representing a single channel image $f: \mathbb{R}^2 \to \mathbb{R}^2$ and $\phi$ a filter $\phi: \mathbb{R}^2 \to \mathbb{R}^2$.
\begin{proof}\label{proof:trans}
\begin{align}
(L_r f \ast \phi)(x) &= \int_{h\in \mathbb{R}^2}f(r^{-1}h)\phi(x - h) dh \\
\intertext{Applying the variable substitution: $h' = r^{-1}h$}
&= \int_{a\in \mathbb{R}^2}f(h')\phi(x - rh') dh' \\
&= \int_{a\in \mathbb{R}^2}f(h')\phi(r(r^{-1}x - h')) dh' \\
&= \int_{a\in \mathbb{R}^2}f(h')\phi(r(r^{-1}x - h')) dh' \\
&= \int_{a\in \mathbb{R}^2}f(h')L_{r^{-1}}\phi(r^{-1}x - h') dh' \\
&= \int_{a\in \mathbb{R}^2}f(h')\phi((x - t) - h') dh' \\
&= L_r (f \ast L_{r^{-1}}\phi)(x) 
\end{align}
\noindent In other words, translational convolution is not equivariant to rotations.
\end{proof}
\noindent Given the two examples above, we now have the proper tools to prove CNN's equivariance property to some arbitrary transformation.

\chapter{Motion Equivariant Networks for Event Cameras with the Temporal Normalization Transform}
\label{content}
\section{Introduction}

Different from single frame images, event-based camera data are parameterized by their spatial coordinates and timestamps. This unique property poses a challenge in learning-based tasks such as object recognition. If we directly take the events as volumes in the spatiotemporal domain, the shape of the events will depend on both the shape and motion of the object that generates the events. The motion can be a result of either camera movements or  object motions. In both cases, the classification network needs to properly handle the different event patterns caused by these motions. Deep learning networks overcome this problem by augmenting the training set with a variety of motions. The networks ``memorizes" different kinds of motions and uses the information to assist it during inference. However, data augmentation would be very costly if a wide spectrum of motions is considered. We propose a novel coordinate transform that normalizes the image coordinates of the events by the timestamp of each event. We show the theoretical proof and the experiments that the transformed events are equivariant under motions in convolution-based deep learning architectures. We construct the N-MOVING-MNIST, a synthetic dataset based on the event-based N-MNIST dataset with additional variations of motions. We evaluate our method against a baseline network with naive volumetric event input. Our method achieves equally well or better performance in all experiments and outperforms the baseline network significantly when there is a larger set of motions in the test set than the training set.
\\\\
Our contributions can be summarized as:
\begin{itemize}
\item The temporal normalization transform (TNT) for events, which transforms events into a space that is equivariant to changes in optical flow for convolutions in a CNN.
\item A CNN architecture which combines a landmark regression network with the TNT to produce representations that are invariant to translation and equivariant to optical flow.
\item The N-MOVING-MNIST dataset, consisting of simulated event data over MNIST images, with many more (30) motion directions than past datasets.
\item Quantitative evaluations on both real and simulated event based datasets, including tests with few motions at training and many different motions at test time.
\end{itemize}

\section{Related Work}
Due to the high speed and dynamic range properties of event cameras, a number of works have attempted to represent the event stream in a form suitable for traditional CNNs for both classification and regression tasks. \cite{moeys2016steering, amir2017low, maqueda2018event, iacono2018towards} generate event histograms, consisting of the number of events at each pixel, and use these as images to classify the position of a robot, perform gesture recognition, estimate steering angle, and perform object classification, respectively. 
\\\\
Several methods have also incorporated the event timestamps in the inputs. \cite{Zhu-RSS-18} represent the events with event histograms as well as the last timestamp at each pixel, to perform self-supervised optical flow estimation. Similarly, \cite{ye2018unsupervised} use the average timestamp at each pixel to perform unsupervised egomotion and depth estimation, and \cite{alonso2018ev} encode the events as a 6 channel image, consisting of positive and negative event histograms, timestamp means and standard deviations, in order to perform semantic segmentation. \cite{zhu2018unsupervised} introduced the discretized event volume, which discretizes the time dimension, and then inserts events using interpolation to perform unsupervised optical flow and egomotion and depth estimation. \cite{lagorce2017hots, sironi2018hats} propose the time surface, which encode the rate of events appearing at each pixel. 
\\\\
In a different vein, \cite{wang2019space} treat the events as a point cloud, and use PointNet~\cite{qi2017pointnet} to process them, while \cite{sekikawa2018constant} propose a solution which learns a set of 2D convolution kernels with associated optical flow directions, which are used to deblur the events at each step of the convolution. 
\\\\
However, these methods either compress the event information into the 2D space, or do not address the issue of equivariance to optical flow when representing events in 3D. However, most current event-based classification datasets are generated with only a limited subset of motions, many from a servo motor with a fixed trajectory. In addition, most datasets have the same motions in the training and test sequences. As a result, the issue of equivariance does not appear, as the network only has to memorize a small number of motions for each class.
\\\\
As mentioned in section \ref{sec:equi-related}, the equivariance for CNNs has been a well studied topic rooted in Lie-generators \cite{ferraro1988relationship, segman1992canonical} and steerability~\cite{freeman1991design, simoncelli1992shiftable, teo1998design}. Recent works like \cite{cohen16_steer_cnns, jacobsen2017dynamic} have extended these ideas for equivariance of CNNs to a number of transformations. New presentations of the data are also proposed to gain automatic equivariance to certain types of transformations in the input space, such as \cite{cohen2018spherical, esteves2018learning}. 
\\\\
Polar Transform Networks~\cite{esteves} obtain equivariance in rotaion and scaling by applying a log polar transform to the images. A spatial transformer is used to maintain the translation invariance property of the CNN. We adopt a similar spatial transformer network to predict a landmark in each image, and apply the temporal normalization transform to obtain invariance to motion from optical flow.
\section{Optical Flow Equivariance for Events}
In this section, we want to explore whether the events are equivariant under the event motion model for translational 2D or 3D convolution. First we define the event generation function, a mapping from the coordinate of an event in the spatiotemporal domain to an event value. We define event value as:
\begin{align}
E :& (\mathbf{x}, t)\rightarrow \{-1, 0, 1\}\\
E(\mathbf{x}_i, t_i)=&\left\{\begin{array}{cc}p_i & \text{if an event was triggered at }(\mathbf{x}_i,t_i)\\ 0 & \text{otherwise} \end{array}\right.
\end{align}
Non-zero event values indicate triggered events with the sign of the values as the polarity of the events. Zero event value means there is not event at this particular $(x, y, t)$. Here is our proof that the flow motion model in \ref{event-motion} is not equivariant to 2D or 3D motion. The proof is an excerpt from our work \cite{zhu2019motion}:

\begin{prop}\label{prop:1}
The optical flow motion model $L_{OF}$ is not equivariant to 2D or 3D convolutions.
\end{prop}

\begin{proof}
For $L_{OF}$ to be equivariant to 2D or 3D convolutions, the following must be true:
\begin{align}
((L_{OF}E)\ast\phi)(\mathbf{x},t)=&L_{OF}(E\ast\phi)(\mathbf{x},t)
\end{align}
Expanding the LHS:\footnote{Here we use the equation for correlation instead of convolution. The proof holds true for both cases, but correlation is the standard form used in many deep learning frameworks.}
\begin{align}
((&L_{OF}E)\ast \phi)(\mathbf{x}, t)\nonumber\\
=&\int\limits_{\mathbf{\xi}\in\mathbb{R}^2, \tau\in\mathbb{R}}L_{OF}E(\mathbf{\xi}, \tau)\phi(\mathbf{\xi}-\mathbf{x}, \tau-t)d\mathbf{\xi}d\tau\\
=&\int\limits_{\mathbf{\xi}\in\mathbb{R}^2, \tau\in\mathbb{R}} E(\mathbf{\xi}+\dot{\mathbf{x}}\tau, \tau)\phi(\mathbf{\xi}-\mathbf{x}, \tau-t) d\mathbf{\xi}d\tau
\intertext{Applying the variable substitution: $\tau'=\tau$, $\mathbf{\xi}':=\mathbf{\xi}+\dot{\mathbf{x}}\tau'$.}
d\mathbf{\xi}&d\tau=d\mathbf{\xi}'d\tau'\\
((&L_{OF}E)\ast \phi)(\mathbf{x}, t)\nonumber\\
=&\int\limits_{\substack{\mathbf{\xi}'\in\mathbb{R}^2\\\tau'\in\mathbb{R}}}E(\mathbf{\xi}', \tau')\phi(\mathbf{\xi}'-(\mathbf{x}+\dot{\mathbf{x}}\tau'), \tau'-t)d\mathbf{\xi}'d\tau' \label{proof:lhs}
\end{align}
We then expand the RHS:
\begin{align}
L&_{OF}((E\ast \phi))(\mathbf{x}, t)\nonumber\\
=&(E\ast\phi)(\mathbf{x}+\dot{\mathbf{x}}t, t)\\
=&\int\limits_{\substack{\mathbf{\xi}'\in\mathbb{R}^2\\\tau'\in\mathbb{R}}}E(\mathbf{\xi}', \tau')\phi(\mathbf{\xi}'-(\mathbf{x}+\dot{\mathbf{x}}t), \tau'-t)d\mathbf{\xi}'d\tau' \label{proof:rhs}
\end{align}
Although \ref{proof:lhs} and \ref{proof:rhs} have very similar form, the major difference is that the LHS \ref{proof:lhs} is integrating over $\tau'$. This means $\phi(\mathbf{\xi}'-(\mathbf{x}+\dot{\mathbf{x}}\tau')$ will have different values depending on the integrand $\tau'$ in the first parameter of the function. On the other hand, RHS \ref{proof:rhs} has constant value $t$ as the parameter of the $\psi$ function. Therefore, the input is not equivariant to 3D convolution under the optical flow transformation.
\\\\
In the 2D case, equivariance is lost as the optical flow transformation is a 3D operation, which cannot be applied to the 2D output activations of the convolution.
\end{proof}

\section{Temporal Normalization Transform (TNT)}

\begin{figure}
\centering
\includegraphics[width=0.4\linewidth]{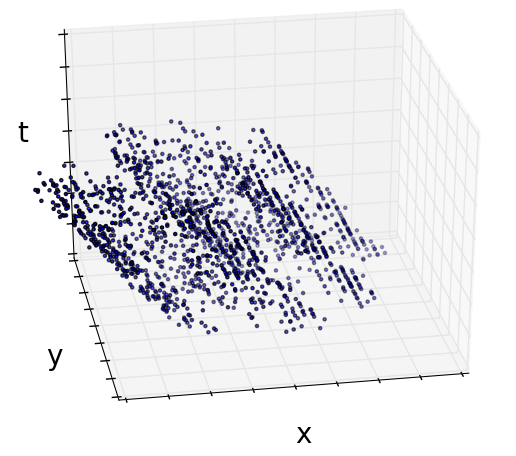}
\includegraphics[width=0.4\linewidth]{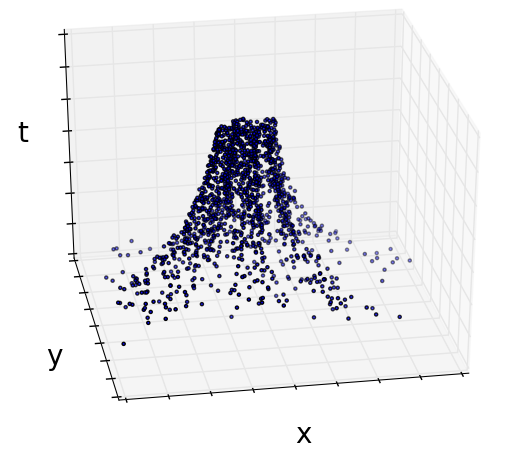}

\includegraphics[width=0.4\linewidth]{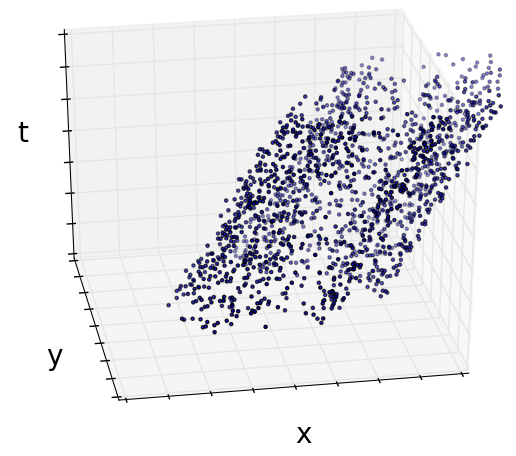}
\includegraphics[width=0.4\linewidth]{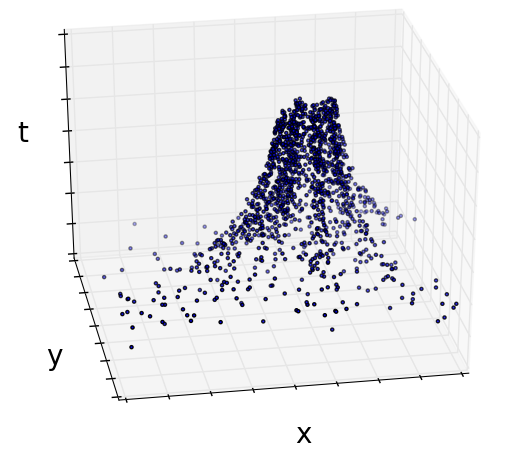}
\caption{Left: Raw input events. Right: Corresponding transformed events.}
\label{fig:transformation}
\end{figure}

We propose the temporal normalization transform $\rho$, a coordinate transform that normalizes the spatial coordinate of the events by their timestamps. As shown in later proofs, the transform maps the events to another domain under which convolution is equivariant to 3D shearing with respect to $t$. Recall that we define the motion model of the events in \ref{event-motion}:
\begin{align*}
L_{OF} \begin{pmatrix}
\mathbf{x_i}\\t_i
\end{pmatrix} = 
\begin{pmatrix}
\mathbf{x_i} + \mathbf{\dot{x}}t_i\\t_i
\end{pmatrix}
\end{align*}
\\
Our idea is to convert the transformation above into a translation in another domain after our transform. The transformation takes an event and scale the pixel position of it by the reciprocal of the timestamps:
\begin{align}
\rho : (\mathbf{x}, t)\rightarrow (\mathbf{x}_\rho, t_\rho)=\left(\frac{\mathbf{x}}{t}, t\right)
\end{align}
Examples of the event volumes and corresponding transformed event volumes can be found in \ref{fig:transformation}. We have two volumes of the same digit moving with different optical flows. The two original event volumes look very different, while the transformed volumes form patterns that are independent of the flow direction. Then we show the proof that the transformed event function is equivariant to 2D and 3D convolution. The proof is an excerpt from our work \cite{zhu2019motion}:

\begin{prop}\label{prop:2}
$\rho(L_{OF})$ is equivariant to both 2D and 3D convolutions. A change in optical flow in $L_{OF}$ is converted to a translation in $\rho(L_{OF})$.
\end{prop}
\begin{proof}
\begin{align}
((L&_{OF}E(\rho))\ast \phi)(\mathbf{x}, t)\nonumber\\
=&\int\limits_{\substack{\mathbf{\xi}_\rho\in\mathbb{R}^2\\\tau_\rho\in\mathbb{R}}} E(\mathbf{\xi}_\rho+\dot{\mathbf{x}}, \tau_\rho)\phi(\xi_\rho-\mathbf{x}, \tau_\rho-t) d\mathbf{\xi}_\rho d\tau_\rho
\intertext{Applying the variable substitution: $\tau':=\tau_\rho, \xi':=\xi_\rho+\dot{\mathbf{x}}$.}
d\xi_\rho&d\tau_\rho=d\xi'd\tau'\\
((L&_{OF}E(\rho))\ast \phi)(\mathbf{x}, t)\nonumber\\
=&\int\limits_{\substack{\mathbf{\xi}'\in\mathbb{R}^2\\\tau'\in\mathbb{R}}} E\left(\mathbf{\xi}', \tau'\right)\phi(\xi'-(\mathbf{x}+\dot{x}), \tau'-t) d\mathbf{\xi}'d\tau'\\
=&L_{OF}(E(\rho)\ast \phi)(\mathbf{x}, t)
\end{align}
\end{proof}
\noindent A similar proof can be written for 2D convolution:
\begin{proof}
\begin{align}
((L&_{OF}E(\rho))\ast \phi)(\mathbf{x})\nonumber\\
=&\int\limits_{\substack{\mathbf{\xi}_\rho\in\mathbb{R}^2\\t \in\mathbb{R}}} E(\mathbf{\xi}_\rho+\dot{\mathbf{x}}, t_\rho)\phi(\xi_\rho-\mathbf{x}) d\mathbf{\xi}_\rho dt_\rho
\intertext{Applying the variable substitution: $t':=t_\rho, \xi':=\xi_\rho+\dot{\mathbf{x}}$.}
d\xi_\rho&dt_\rho=d\xi'dt'\\
((L&_{OF}E(\rho))\ast \phi)(\mathbf{x})\nonumber\\
=&\int\limits_{\substack{\mathbf{\xi}'\in\mathbb{R}^2\\t'\in\mathbb{R}}} E\left(\mathbf{\xi}', \tau'\right)\phi(\xi'-(\mathbf{x}+\dot{x})) d\mathbf{\xi}'dt'\\
=&L_{OF}(E(\rho)\ast \phi)(\mathbf{x})
\end{align}
\end{proof}
\noindent In other words, when we transform the event using TNT, the motion of the events is mapped to a translation that is equal to the optical flow. Intuitively, since images and volumes are equivariant to 2D and 3D convolution under translation, the transformed events will be equivariant to 2D and 3D convolution on a constant motion in the original coordinate system.

\section{Landmark Regressor}
It needs to be noted that, while we show the proof that the transformed events are equivariant to convolution under a motion in the original coordinate system, we lose translation equivariance. We tackle this problem by applying a spatial transformer network to regress a landmark position on the target, following \cite{esteves}. The landmark position is not necessarily the center of the target as long as it is consistent among all the targets. We show proofs that applying our transform after centering the events at a consistent landmark position on the target yields translation invariance to convolution. The proof is an excerpt from our work \cite{zhu2019motion}:

\begin{prop}\label{prop:3}
The motion scaling transformation $\rho$ is translation invariant to convolutions after centering all events around a common landmark on the object. The position of this center is arbitrary as long as it is consistent between objects.
\end{prop}

\begin{proof}
\label{sec:prop3_proof}

Let $\mathbf{s}\in\mathbb{R}^2$ be a translation of the events. Given an accurate landmark regression network, the predicted landmark position is $\mathbf{s}$ translated from the original landmark position. Let a translation of the events be represented by the function $T_\mathbf{k} : (\mathbf{x}, t)\rightarrow(\bar{\mathbf{x}}, \bar{t})=(\mathbf{x}+\mathbf{k}, t)$, $s'$ is the actual translation as predicted by the landmark regressor.
\begin{align}
((\rho& T_{\mathbf{s}}L_{OF}E)\ast \phi)(\mathbf{x}, \tau)\nonumber\\
=&\int\limits_{\mathbf{\xi}\in\mathbb{R}^2, \tau\in\mathbb{R}} E\left(\frac{\mathbf{\xi}+\mathbf{s'}}{\tau}+\dot{\mathbf{x}}, \tau\right)\phi(\mathbf{\xi}-\mathbf{x}, \tau) d\mathbf{\xi}d\tau\\
=&\int\limits_{\mathbf{\xi}_\rho\in\mathbb{R}^2, \tau_\rho\in\mathbb{R}} E\left(\mathbf{\xi}_\rho+\frac{\mathbf{s'}}{\tau_\rho}+\dot{\mathbf{x}}, \tau_\rho\right)\phi(\mathbf{\xi}_\rho-\mathbf{x}, \tau_\rho) d\mathbf{\xi}_\rho d\tau_\rho
\intertext{Applying the variable substitution: $\tau':=\tau$, $\xi':=\xi_\rho+\frac{\mathbf{s'}}{\tau'}+\dot{\mathbf{x}}$.}
=&\int\limits_{\mathbf{\xi}'\in\mathbb{R}^2, \tau'\in\mathbb{R}} E(\xi', \tau')\phi\left(\mathbf{\xi}'-\left(\mathbf{x}+\frac{\mathbf{s'}}{\tau'}+\dot{\mathbf{x}}\right), \tau'\right) d\mathbf{\xi}'d\tau'\\
\intertext{Because we assume an accurate landmark regressor, $s' = s$.}
=&\int\limits_{\mathbf{\xi}'\in\mathbb{R}^2, \tau'\in\mathbb{R}} E(\xi', \tau')\phi\left(\mathbf{\xi}'-\left(\mathbf{x}+\frac{\mathbf{s}}{\tau'}+\dot{\mathbf{x}}\right), \tau'\right) d\mathbf{\xi}'d\tau'\\
=&\rho T_{\mathbf{s}}L_{OF}(E(\rho)\ast \phi)(\mathbf{x}, \tau)
\end{align}
\end{proof}
\noindent Note that in this derivation, we only consider the translation of the events. The location of the landmark within the event volume is not important as long as the translation between any two landmarks of the event volumes is the same as the ground truth translation between two event volumes. This conclusion simplifies the task of the landmark regressor since it no longer has a single pixel position to predict but a set of consistency constraints. Our landmark regressor is composed of three layers of 3 by 3 convolution layers followed by a 1 by 1 convolution layer. The output of the network is a heatmap whose centroid we use to center the object.

\section{Implementation}
\subsection{N-MOVING-MNIST Dataset}
In order to evaluate the performance of our method, we need a dataset that has motions with different sets of magnitude and direction. The N-MNIST and N-CALTECH101 datasets \cite{orchard2015converting} are two datasets that convert the standard MNIST and CALTECH101 datasets into events, captured by an event camera moving in parallel to a screen displaying various  objects. However, the existing datasets lack diversity in movements. With a small set of motions, the neural neworks can memorize the types of movements in the training set. Therefore, it would be hard to test the generality of the method by directly using these datasets.
\\\\
Therefore, we construct a new dataset that has much more variation in object motion. The dataset is generated using the Event Camera Simulator \cite{mueggler2017event}. The simulator takes in a sequence of images with certain frame rate and produces an event stream. We take the images from the original MNIST dataset and simulate 30 sets of events, each with a different motion. The direction and the motion of the events are controlled by designing a trajectory of the camera with respect to a static digit in the scene. In this way, we have a dataset with 30 sets of different motions. We rendered the dataset in a distributed manner on 24 CPU cores and it took around 4 days for it to complete.
\subsection{Network Architecture}
In order to compare the effectiveness of our method, we use two identical networks with and without the transform. The networks are designed to be small because we want the network to have insufficient capacity to memorize different types of motions. We construct a small CNN consisting of two convolutional layers with ReLU activations. Each layer is followed by average pooling. In the end, we have two fully connected layers that map the features from convolution to N outputs as an one-hot vector indicating the probability of each class prediction. 

\begin{figure*}[t]
\centering
\includegraphics[width=\linewidth]{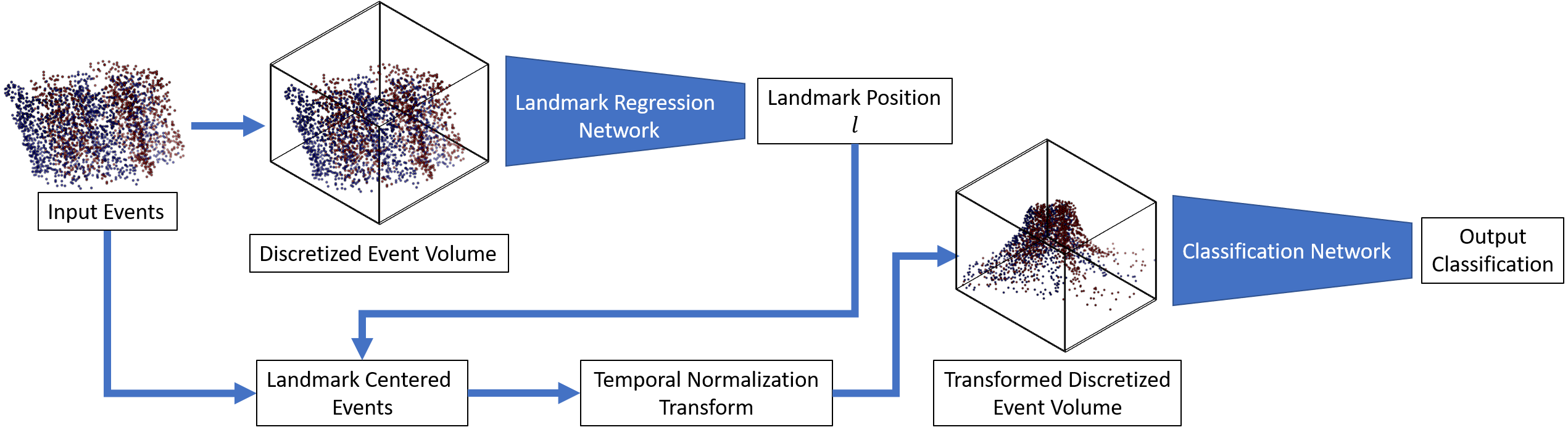}
\caption{Overview of the proposed pipeline. The input events are first converted into a discretized event volume, and passed through the landmark regression network to estimate the landmark position, $l$. This is used to center the events around $l$, on which the temporal normalization transform is applied. A second discretized event volume is generated on the transformed, centered events, and finally passed through the classification network to generate the final output classification.}
\label{fig:pipeline}
\end{figure*}
\subsection{Pipeline}
The pipeline of our implementation is shown in Figure \ref{fig:pipeline}. Our pipeline has three main parts: the landmark regressor, Temporal Normalization Transform and the classification network. The input volume first goes through the landmark regressor network. The regressor generates a single landmark position and centers the event volume by it. The event volume then goes through our Temporal Normalization Transform layer so that all the events are mapped to the new coordinate space. The events are then represented as interpolated discretized event images, as described in \ref{event-represent}. The event images are then passed into the final classification network. The training process is fully end-to-end. Similarly to the Spatial Transformer Network \cite{jaderberg2015spatial}, the loss function of the network is simply the classification loss. The learning process of the regressor is not supervised by other losses. 

\subsection{Evaluation}
In all experiments, we compare our method with the a baseline CNN network. Both methods take in as an input a series of events, preprocess the events to their separate volume forms and feed them into the neural network. Our method performs TNT on the data before converting them to discretized event volumes whereas the baseline pipeline directly converts the raw events int discretized a event volume. The N-MNIST dataset is collected using a moving camera in front of a computer screen. Each N-MNIST digit has three camera motions and we segment each motion out by thresholding on the timestamp. In our evaluation, we train and test on combinations of the N-MNIST dataset and our MOVING-N-MNIST dataset. We also perform an ablation study on the landmark regression network (TNT+regress) with a heuristic that centers events around the center of the image (TNT). To evaluate the method's ability to generalize to unseen motions, we design the experiments to fit different scenarios: same motions in training and test, one motion in training and all motions in test. In addition, we use simulated data and real dataset in both test and train to evaluate the performance of the method on new datasets with unseen motions.

\subsection{Implementation Details}
All models are trained for 60,000 iterations with a batch size of 64, and saturated validation accuracy before stopping. When training with the landmark regressor, random translations are applied as data augmentation.
\\\\
One issue for implementation is the tendency for the transformed coordinates to grow towards infinity as $t\rightarrow 0$. Due to the need to discretize the spatial dimension at each convolution layer, it is prohibitively expensive to try to encompass all transformed events when discretizing. Instead, we have chosen to omit any points that fall outside a predefined image boundary, $[W, H]$. For this work, we have kept the same transformed image size as the original input image. In addition, the number of events falling out of the image can be controlled by scaling the timestamps before applying the transform, as equivariance is maintained for any constant scaling of the timestamps. However, there is a tradeoff between minimizing the number of events leaving the image and minimizing the compression of events at the highest timestamps. Due to the discretization in the spatial domain, transformed events which are very close together will be placed in the same voxel of the volume. In practice, we found that scaling the timestamps to be between 0 and $B-1$, where $B=9$ is the number of bins, works well. With this scaling, only events in the first bin may be transformed out of the image. 

\subsection{Results}
The results from all the experiments can be found in Table \ref{tab:results_table}. All the models perform very well when trained and tested on all of N-MNIST dataset. This is expected since the network has seen and memorized the limited set of motions. When we train the model with only one type of motion and tests the model on all types of motions, the advantage of our TNT starts to show. Since we are only tested on one set of optical flow, the baseline network struggles to recognize a digit that moves in another direction in the scene. On the other hand, because of the equivariance property of our transformed events to 2D convolution, the features of the network are translated by the optical flow caused by the motion. Imagine all the events are translated version of each other by their difference in motion. CNN is known for recognizing translated objects well and the transformation makes the CNN robust to different types of motions. The tables shows that our proposed transform improves the performance of the baseline network in all cases when we have significantly more variation in motion in the test set than in the training set.  

\begin{table}[t]
\centering
\begin{tabular}{c|ccccc} 
train/test sets & all/all & 1/all & 1/train & all/sim & 1/sim \\\hline
Baseline & \textbf{0.991} & 0.437 & 0.442 & 0.396 & 0.207 \\
TNT & 0.981 & 0.468 & 0.464 & \textbf{0.592} & 0.318\\
TNT+regress & 0.981 & \textbf{0.485} & \textbf{0.481} & 0.566 & \textbf{0.324}
\end{tabular}
\caption{Results from experiments on N-MNIST and N-MOVING-MNIST. Experiments are denoted by train$\_$set/test$\_$set, with the following labels: 1 - trained on the first motion in the N-MNIST train set. all - training/testing on all 3 motions in the N-MNIST train/test set respectively. train - testing on all 3 motions in the N-MNIST train set. sim - testing on all motions in the N-MOVING-MNIST test set.}
\label{tab:results_table}
\end{table}

\subsection{Discussion}
In our experiments, we explore the hard case when we test on motions unseen in the training set. To test the generality of our method, we will have to intentionally introduce motion variation into testing. This is also a potential issue that we recognize in the evaluation of object classification works for events. The methods are often evaluated on different objects of the same class of the same set of motions. If we only look at the first column of Table \ref{tab:results_table}, it is easy to assume that the network will generalize well to real-life scenes. However, given the small temporal window typically used in processing event data, a slight camera shake can produce volumes with multiple different movements. This is well-handled in CNN on images, since images are invariant to 2D convolution under translation. For event-based camera data, the change can significantly increase the difficulty of matching a learned object template.
\\\\
A potential solution would be collecting datasets with a diverse set of motions and let the network remember these motions. However, it would be very expensive to generate event-based camera data that vary in both objects, deformation and motion. The fact that we cannot easily obtain complete set of motions makes it valuable to explore the invariance within the data. With our proposed method, we are able to generalize to unseen data more effectively without the use of extensive data augmentation.
\\\\However, there is still potential improvement to be made to boost the performance of our method. First, we need to look at a few factors that affect the performance of classification neural networks exclusive to event data:
\begin{itemize}
\item For the event data, we only capture the relative change of log light intensity. As stated above, the motions will deform the shape of the event volumes because the events will be move spatially at the flow direction as time changes. However, this is not the only way motion changes the shape of the events. An event volume will also vary its shape based on the combination of the motion and the image gradient. For example, if an edge moves in the direction orthogonal to the image gradient of a pixel, there will be no events generated for the pixel. This means there are additional challenges for the network to recognize the same object when some parts of it are missing or different due to an unseen motion.
\item We also need to consider the information embedded in the polarities. The polarity of an event indicates the direction of movement if we fix all other aspects of the same object. In other words, reversing the direction of motion will produce a different polarity on the same pixel. While it is possible to ignore the polarities and use them as point clouds, it would be also useful to explicitly exploit the change of polarity to help with understanding the motion. As of now, we rely on the networks to learn the information automatically.

\end{itemize}

\chapter{Conclusions}

Event-based camera is a novel vision sensor that carries good potential in achieving better performance in certain tasks than traditional cameras. While the camera has advantages in latency, power consumption and dynamic range, the classical algorithms in computer vision do not suitably apply to event data in many cases. The representation and processing of event camera data still remain an active area of research. Our recent project, along with its theoretical background, is included to show how to manipulate the representation of event data to improve the robustness of a classification network to motion. While we restrict this work to object classification, there are many potential applications of the method. The deformation of event volumes due to motion is common in most tasks that require pattern matching. Through this work, we hope to bring a better understanding of both the event camera and the mathematical basis of equivariance. In the next section Future Work, we will discuss the potential methods to further improve the generality of our method.
\section{Future Work}
Our Temporal Normalization Network aims to improve robustness of event-based camera algorithms to variation in motions. While our work has shown to improve the performance of existing architecture, we make the global optical flow assumption. Although the constraint has been used extensively and shown to be reasonable, it is not necessarily valid if the object of interest is non-rigid or the temporal window is longer. In future work, we plan to loosen the constraint and have local consistency in flow. This means that translation equivaraince is no longer learned through a land mark regressor. Instead, new transforms or kernel functions can be developed to achieve equivaraince in both motion and translation. Similarly to \cite{worrall16_harmon_networ}, a constraint on the filter weight space can developed to obtain the motion equivariance by directly baking equivariance into the CNN architecture.
\bibliographystyle{plainnat}
\bibliography{ref.bib}

\begin{thebibliography}{46}
\providecommand{\natexlab}[1]{#1}
\providecommand{\url}[1]{\texttt{#1}}
\expandafter\ifx\csname urlstyle\endcsname\relax
  \providecommand{\doi}[1]{doi: #1}\else
  \providecommand{\doi}{doi: \begingroup \urlstyle{rm}\Url}\fi

\bibitem[Alonso and Murillo(2018)]{alonso2018ev}
I{\~n}igo Alonso and Ana~C Murillo.
\newblock Ev-segnet: Semantic segmentation for event-based cameras.
\newblock \emph{arXiv preprint arXiv:1811.12039}, 2018.

\bibitem[Amir et~al.()Amir, Taba, Berg, Melano, McKinstry, Di~Nolfo, Nayak,
  Andreopoulos, Garreau, Mendoza, et~al.]{amir2017low}
Arnon Amir, Brian Taba, David Berg, Timothy Melano, Jeffrey McKinstry, Carmelo
  Di~Nolfo, Tapan Nayak, Alexander Andreopoulos, Guillaume Garreau, Marcela
  Mendoza, et~al.
\newblock A low power, fully event-based gesture recognition system.

\bibitem[Cohen and Welling(2016{\natexlab{a}})]{cohen16_steer_cnns}
Taco~S. Cohen and Max Welling.
\newblock Steerable cnns.
\newblock 2016{\natexlab{a}}.
\newblock URL \url{http://arxiv.org/abs/1612.08498v1}.

\bibitem[Cohen and Welling(2016{\natexlab{b}})]{cohen2016group}
Taco~S Cohen and Max Welling.
\newblock Group equivariant convolutional networks.
\newblock \emph{arXiv preprint arXiv:1602.07576}, 2016{\natexlab{b}}.

\bibitem[Cohen et~al.(2018)Cohen, Geiger, K{\"o}hler, and
  Welling]{cohen2018spherical}
Taco~S Cohen, Mario Geiger, Jonas K{\"o}hler, and Max Welling.
\newblock Spherical cnns.
\newblock \emph{arXiv preprint arXiv:1801.10130}, 2018.

\bibitem[Cook et~al.(2011)Cook, Gugelmann, Jug, Krautz, and
  Steger]{cook2011interacting}
Matthew Cook, Luca Gugelmann, Florian Jug, Christoph Krautz, and Angelika
  Steger.
\newblock Interacting maps for fast visual interpretation.
\newblock In \emph{The 2011 International Joint Conference on Neural Networks},
  pages 770--776. IEEE, 2011.

\bibitem[Dalal and Triggs(2005)]{dalal2005histograms}
Navneet Dalal and Bill Triggs.
\newblock Histograms of oriented gradients for human detection.
\newblock In \emph{international Conference on computer vision \& Pattern
  Recognition (CVPR'05)}, volume~1, pages 886--893. IEEE Computer Society,
  2005.

\bibitem[Esteves et~al.(2017)Esteves, Allen-Blanchette, Zhou, and
  Daniilidis]{esteves}
Carlos Esteves, Christine Allen-Blanchette, Xiaowei Zhou, and Kostas
  Daniilidis.
\newblock Polar transformer networks.
\newblock \emph{arXiv preprint arXiv:1709.01889}, 2017.

\bibitem[Esteves et~al.(2018)Esteves, Allen-Blanchette, Makadia, and
  Daniilidis]{esteves2018learning}
Carlos Esteves, Christine Allen-Blanchette, Ameesh Makadia, and Kostas
  Daniilidis.
\newblock Learning so (3) equivariant representations with spherical cnns.
\newblock In \emph{European Conference on Computer Vision}, pages 54--70.
  Springer, 2018.

\bibitem[Ferraro and Caelli(1988)]{ferraro1988relationship}
Mario Ferraro and Terry~M Caelli.
\newblock Relationship between integral transform invariances and lie group
  theory.
\newblock \emph{JOSA A}, 5\penalty0 (5):\penalty0 738--742, 1988.

\bibitem[Freeman et~al.(1991)Freeman, Adelson, et~al.]{freeman1991design}
William~T Freeman, Edward~H Adelson, et~al.
\newblock The design and use of steerable filters.
\newblock \emph{IEEE Transactions on Pattern analysis and machine
  intelligence}, 13\penalty0 (9):\penalty0 891--906, 1991.

\bibitem[Gallego et~al.(2018)Gallego, Rebecq, and
  Scaramuzza]{gallego2018unifying}
Guillermo Gallego, Henri Rebecq, and Davide Scaramuzza.
\newblock A unifying contrast maximization framework for event cameras, with
  applications to motion, depth, and optical flow estimation.
\newblock In \emph{Proceedings of the IEEE Conference on Computer Vision and
  Pattern Recognition}, pages 3867--3876, 2018.

\bibitem[Gehrig et~al.(2018)Gehrig, Rebecq, Gallego, and
  Scaramuzza]{gehrig2018asynchronous}
Daniel Gehrig, Henri Rebecq, Guillermo Gallego, and Davide Scaramuzza.
\newblock Asynchronous, photometric feature tracking using events and frames.
\newblock In \emph{Proceedings of the European Conference on Computer Vision
  (ECCV)}, pages 750--765, 2018.

\bibitem[Iacono et~al.(2018)Iacono, Weber, Glover, and
  Bartolozzi]{iacono2018towards}
Massimiliano Iacono, Stefan Weber, Arren Glover, and Chiara Bartolozzi.
\newblock Towards event-driven object detection with off-the-shelf deep
  learning.
\newblock In \emph{2018 IEEE/RSJ International Conference on Intelligent Robots
  and Systems (IROS)}, pages 1--9. IEEE, 2018.

\bibitem[Jacobsen et~al.(2017)Jacobsen, De~Brabandere, and
  Smeulders]{jacobsen2017dynamic}
J{\"o}rn-Henrik Jacobsen, Bert De~Brabandere, and Arnold~WM Smeulders.
\newblock Dynamic steerable blocks in deep residual networks.
\newblock \emph{arXiv preprint arXiv:1706.00598}, 2017.

\bibitem[Jaderberg et~al.(2015)Jaderberg, Simonyan, Zisserman,
  et~al.]{jaderberg2015spatial}
Max Jaderberg, Karen Simonyan, Andrew Zisserman, et~al.
\newblock Spatial transformer networks.
\newblock In \emph{Advances in neural information processing systems}, pages
  2017--2025, 2015.

\bibitem[Kim et~al.(2008)Kim, Handa, Benosman, Ieng, and
  Davison]{kim2008simultaneous}
Hanme Kim, Ankur Handa, Ryad Benosman, Sio-Hoi Ieng, and Andrew~J Davison.
\newblock Simultaneous mosaicing and tracking with an event camera.
\newblock \emph{J. Solid State Circ}, 43:\penalty0 566--576, 2008.

\bibitem[Kim et~al.(2016)Kim, Leutenegger, and Davison]{kim2016real}
Hanme Kim, Stefan Leutenegger, and Andrew~J Davison.
\newblock Real-time 3d reconstruction and 6-dof tracking with an event camera.
\newblock In \emph{European Conference on Computer Vision}, pages 349--364.
  Springer, 2016.

\bibitem[Kleinfelder et~al.(2001)Kleinfelder, Lim, Liu, and
  El~Gamal]{kleinfelder200110000}
Stuart Kleinfelder, SukHwan Lim, Xinqiao Liu, and Abbas El~Gamal.
\newblock A 10000 frames/s cmos digital pixel sensor.
\newblock \emph{IEEE Journal of Solid-State Circuits}, 36\penalty0
  (12):\penalty0 2049--2059, 2001.

\bibitem[Kyatkin and Chirikjian(2000)]{kyatkin2000algorithms}
Alexander~B Kyatkin and Gregory~S Chirikjian.
\newblock Algorithms for fast convolutions on motion groups.
\newblock \emph{Applied and Computational Harmonic Analysis}, 9\penalty0
  (2):\penalty0 220--241, 2000.

\bibitem[Lagorce et~al.(2017)Lagorce, Orchard, Galluppi, Shi, and
  Benosman]{lagorce2017hots}
Xavier Lagorce, Garrick Orchard, Francesco Galluppi, Bertram~E Shi, and Ryad~B
  Benosman.
\newblock Hots: a hierarchy of event-based time-surfaces for pattern
  recognition.
\newblock \emph{IEEE transactions on pattern analysis and machine
  intelligence}, 39\penalty0 (7):\penalty0 1346--1359, 2017.

\bibitem[Lee et~al.(2016)Lee, Delbruck, and Pfeiffer]{lee2016training}
Jun~Haeng Lee, Tobi Delbruck, and Michael Pfeiffer.
\newblock Training deep spiking neural networks using backpropagation.
\newblock \emph{Frontiers in neuroscience}, 10:\penalty0 508, 2016.

\bibitem[Lichtsteiner et~al.(2008)Lichtsteiner, Posch, and
  Delbruck]{lichtsteiner}
Patrick Lichtsteiner, Christoph Posch, and Tobi Delbruck.
\newblock A 128$\times$128 120 db 15$\mu$ s latency asynchronous temporal
  contrast vision sensor.
\newblock \emph{IEEE journal of solid-state circuits}, 43\penalty0
  (2):\penalty0 566--576, 2008.

\bibitem[Lowe(2004)]{lowe2004distinctive}
David~G Lowe.
\newblock Distinctive image features from scale-invariant keypoints.
\newblock \emph{International journal of computer vision}, 60\penalty0
  (2):\penalty0 91--110, 2004.

\bibitem[Lucas et~al.(1981)Lucas, Kanade, et~al.]{lucas1981iterative}
Bruce~D Lucas, Takeo Kanade, et~al.
\newblock An iterative image registration technique with an application to
  stereo vision.
\newblock 1981.

\bibitem[Maqueda et~al.(2018)Maqueda, Loquercio, Gallego, Garc{\'\i}a, and
  Scaramuzza]{maqueda2018event}
Ana~I Maqueda, Antonio Loquercio, Guillermo Gallego, Narciso Garc{\'\i}a, and
  Davide Scaramuzza.
\newblock Event-based vision meets deep learning on steering prediction for
  self-driving cars.
\newblock In \emph{Proceedings of the IEEE Conference on Computer Vision and
  Pattern Recognition}, pages 5419--5427, 2018.

\bibitem[Moeys et~al.(2016)Moeys, Corradi, Kerr, Vance, Das, Neil, Kerr, and
  Delbr{\"u}ck]{moeys2016steering}
Diederik~Paul Moeys, Federico Corradi, Emmett Kerr, Philip Vance, Gautham Das,
  Daniel Neil, Dermot Kerr, and Tobi Delbr{\"u}ck.
\newblock Steering a predator robot using a mixed frame/event-driven
  convolutional neural network.
\newblock In \emph{Event-based Control, Communication, and Signal Processing
  (EBCCSP), 2016 Second International Conference on}, pages 1--8. IEEE, 2016.

\bibitem[Mueggler et~al.(2017)Mueggler, Rebecq, Gallego, Delbruck, and
  Scaramuzza]{mueggler2017event}
Elias Mueggler, Henri Rebecq, Guillermo Gallego, Tobi Delbruck, and Davide
  Scaramuzza.
\newblock The event-camera dataset and simulator: Event-based data for pose
  estimation, visual odometry, and slam.
\newblock \emph{The International Journal of Robotics Research}, 36\penalty0
  (2):\penalty0 142--149, 2017.

\bibitem[Nguyen et~al.(2017)Nguyen, Do, Caldwell, and
  Tsagarakis]{nguyen2017real}
Anh Nguyen, Thanh-Toan Do, Darwin~G Caldwell, and Nikos~G Tsagarakis.
\newblock Real-time 6dof pose relocalization for event cameras with stacked
  spatial lstm networks.
\newblock \emph{arXiv preprint arXiv:1708.09011}, 2017.

\bibitem[Orchard et~al.(2015{\natexlab{a}})Orchard, Jayawant, Cohen, and
  Thakor]{orchard2015converting}
Garrick Orchard, Ajinkya Jayawant, Gregory~K Cohen, and Nitish Thakor.
\newblock Converting static image datasets to spiking neuromorphic datasets
  using saccades.
\newblock \emph{Frontiers in neuroscience}, 9:\penalty0 437,
  2015{\natexlab{a}}.

\bibitem[Orchard et~al.(2015{\natexlab{b}})Orchard, Meyer, Etienne-Cummings,
  Posch, Thakor, and Benosman]{orchard2015hfirst}
Garrick Orchard, Cedric Meyer, Ralph Etienne-Cummings, Christoph Posch, Nitish
  Thakor, and Ryad Benosman.
\newblock Hfirst: a temporal approach to object recognition.
\newblock \emph{IEEE transactions on pattern analysis and machine
  intelligence}, 37\penalty0 (10):\penalty0 2028--2040, 2015{\natexlab{b}}.

\bibitem[P{\'e}rez-Carrasco et~al.(2013)P{\'e}rez-Carrasco, Zhao, Serrano,
  Acha, Serrano-Gotarredona, Chen, and Linares-Barranco]{perez2013mapping}
Jos{\'e}~Antonio P{\'e}rez-Carrasco, Bo~Zhao, Carmen Serrano, Begona Acha,
  Teresa Serrano-Gotarredona, Shouchun Chen, and Bernab{\'e} Linares-Barranco.
\newblock Mapping from frame-driven to frame-free event-driven vision systems
  by low-rate rate coding and coincidence processing--application to
  feedforward convnets.
\newblock \emph{IEEE transactions on pattern analysis and machine
  intelligence}, 35\penalty0 (11):\penalty0 2706--2719, 2013.

\bibitem[Qi et~al.(2017)Qi, Su, Mo, and Guibas]{qi2017pointnet}
Charles~R Qi, Hao Su, Kaichun Mo, and Leonidas~J Guibas.
\newblock Pointnet: Deep learning on point sets for 3d classification and
  segmentation.
\newblock \emph{Proc. Computer Vision and Pattern Recognition (CVPR), IEEE},
  1\penalty0 (2):\penalty0 4, 2017.

\bibitem[Segman et~al.(1992)Segman, Rubinstein, and Zeevi]{segman1992canonical}
Joseph Segman, Jacob Rubinstein, and Yehoshua~Y Zeevi.
\newblock The canonical coordinates method for pattern deformation: Theoretical
  and computational considerations.
\newblock \emph{IEEE Transactions on Pattern Analysis and Machine
  Intelligence}, 14\penalty0 (12):\penalty0 1171--1183, 1992.

\bibitem[Sekikawa et~al.(2018)Sekikawa, Ishikawa, Hara, Yoshida, Suzuki, Sato,
  and Saito]{sekikawa2018constant}
Yusuke Sekikawa, Kohta Ishikawa, Kosuke Hara, Yuuichi Yoshida, Koichiro Suzuki,
  Ikuro Sato, and Hideo Saito.
\newblock Constant velocity 3d convolution.
\newblock In \emph{2018 International Conference on 3D Vision (3DV)}, pages
  343--351. IEEE, 2018.

\bibitem[Simoncelli et~al.(1992)Simoncelli, Freeman, Adelson, and
  Heeger]{simoncelli1992shiftable}
Eero~P Simoncelli, William~T Freeman, Edward~H Adelson, and David~J Heeger.
\newblock Shiftable multiscale transforms.
\newblock \emph{IEEE transactions on Information Theory}, 38\penalty0
  (2):\penalty0 587--607, 1992.

\bibitem[Sironi et~al.()Sironi, Brambilla, Bourdis, Lagorce, and
  Benosman]{sironi2018hats}
Amos Sironi, Manuele Brambilla, Nicolas Bourdis, Xavier Lagorce, and Ryad
  Benosman.
\newblock Hats: Histograms of averaged time surfaces for robust event-based
  object classification.

\bibitem[Teo and Hel-Or(1998)]{teo1998design}
Patrick~C Teo and Yacov Hel-Or.
\newblock Design of multi-parameter steerable functions using cascade basis
  reduction.
\newblock In \emph{Computer Vision, 1998. Sixth International Conference on},
  pages 187--192. IEEE, 1998.

\bibitem[Wang et~al.(2019)Wang, Zhang, Yuan, and Lu]{wang2019space}
Qinyi Wang, Yexin Zhang, Junsong Yuan, and Yilong Lu.
\newblock Space-time event clouds for gesture recognition: from rgb cameras to
  event cameras.
\newblock \emph{IEEE Winter Conference on Applications of Computer Vision},
  2019.

\bibitem[Worrall et~al.(2016)Worrall, Garbin, Turmukhambetov, and
  Brostow]{worrall16_harmon_networ}
Daniel~E Worrall, Stephan~J Garbin, Daniyar Turmukhambetov, and Gabriel~J
  Brostow.
\newblock Harmonic networks: Deep translation and rotation equivariance.
\newblock \emph{arXiv preprint arXiv:1612.04642}, 2016.

\bibitem[Ye et~al.(2018)Ye, Mitrokhin, Parameshwara, Ferm{\"u}ller, Yorke, and
  Aloimonos]{ye2018unsupervised}
Chengxi Ye, Anton Mitrokhin, Chethan Parameshwara, Cornelia Ferm{\"u}ller,
  James~A Yorke, and Yiannis Aloimonos.
\newblock Unsupervised learning of dense optical flow and depth from sparse
  event data.
\newblock \emph{arXiv preprint arXiv:1809.08625}, 2018.

\bibitem[Zhao et~al.(2015)Zhao, Ding, Chen, Linares-Barranco, and
  Tang]{zhao2015feedforward}
Bo~Zhao, Ruoxi Ding, Shoushun Chen, Bernabe Linares-Barranco, and Huajin Tang.
\newblock Feedforward categorization on aer motion events using cortex-like
  features in a spiking neural network.
\newblock \emph{IEEE transactions on neural networks and learning systems},
  26\penalty0 (9):\penalty0 1963--1978, 2015.

\bibitem[Zhu et~al.(2018{\natexlab{a}})Zhu, Yuan, Chaney, and
  Daniilidis]{Zhu-RSS-18}
Alex Zhu, Liangzhe Yuan, Kenneth Chaney, and Kostas Daniilidis.
\newblock Ev-flownet: Self-supervised optical flow estimation for event-based
  cameras.
\newblock In \emph{Proceedings of Robotics: Science and Systems}, Pittsburgh,
  Pennsylvania, June 2018{\natexlab{a}}.
\newblock \doi{10.15607/RSS.2018.XIV.062}.

\bibitem[Zhu et~al.(2017)Zhu, Atanasov, and Daniilidis]{zhu2017event}
Alex~Zihao Zhu, Nikolay Atanasov, and Kostas Daniilidis.
\newblock Event-based feature tracking with probabilistic data association.
\newblock In \emph{2017 IEEE International Conference on Robotics and
  Automation (ICRA)}, pages 4465--4470. IEEE, 2017.

\bibitem[Zhu et~al.(2018{\natexlab{b}})Zhu, Yuan, Chaney, and
  Daniilidis]{zhu2018unsupervised}
Alex~Zihao Zhu, Liangzhe Yuan, Kenneth Chaney, and Kostas Daniilidis.
\newblock Unsupervised event-based learning of optical flow, depth, and
  egomotion.
\newblock \emph{arXiv preprint arXiv:1812.08156}, 2018{\natexlab{b}}.

\bibitem[Zhu et~al.(2019)Zhu, Wang, and Daniilidis]{zhu2019motion}
Alex~Zihao Zhu, Ziyun Wang, and Kostas Daniilidis.
\newblock Motion equivariant networks for event cameras with the temporal
  normalization transform.
\newblock \emph{arXiv preprint arXiv:1902.06820}, 2019.

\end{thebibliography}
\end{document}